\documentclass[11pt]{article}

\usepackage[a4paper,margin=1in]{geometry}

\usepackage[T1]{fontenc}
\usepackage[utf8]{inputenc}
\usepackage[english]{babel}
\usepackage{amsmath,amssymb}
\usepackage{newtxtext,newtxmath}   
\usepackage{bm}
\usepackage{microtype}

\usepackage{graphicx}
\usepackage{booktabs}
\usepackage{enumitem}
\usepackage{float}
\floatstyle{ruled}
\newfloat{algorithm}{tbp}{loa}
\floatname{algorithm}{Algorithm}
\usepackage[font=small,labelfont=bf,labelsep=period,skip=6pt]{caption}

\usepackage{titlesec}
\titleformat{\section}{\large\bfseries}{\thesection}{0.6em}{}
\titleformat{\subsection}{\normalsize\bfseries}{\thesubsection}{0.6em}{}
\titleformat{\subsubsection}{\normalsize\itshape}{\thesubsubsection}{0.6em}{}
\titlespacing*{\section}{0pt}{14pt plus 3pt minus 2pt}{6pt}
\titlespacing*{\subsection}{0pt}{11pt plus 2pt minus 2pt}{4pt}
\titlespacing*{\subsubsection}{0pt}{10pt plus 2pt minus 2pt}{3pt}

\usepackage{fancyhdr}
\usepackage[numbers,sort&compress]{natbib}
\usepackage{xcolor}
\definecolor{linkcolor}{HTML}{16436E}
\usepackage{hyperref}
\hypersetup{
  colorlinks=true,
  linkcolor=linkcolor,
  citecolor=linkcolor,
  urlcolor=linkcolor,
  pdftitle={Embedded Graph Flows for Categorical Graph Generation},
  pdfauthor={Ethan Ma, Zihan Wang, Chris Siu Yeung Chow, Xinguo Feng, Qingqing Li, Rui Jiang, Naipeng Dong, Guangdong Bai},
}
\newcommand{\affil}[2]{\textsuperscript{#1}\,#2}

\renewenvironment{abstract}{%
  \noindent\rule{\textwidth}{0.4pt}\par\vspace{6pt}%
  \centerline{\normalsize\bfseries\abstractname}\par\vspace{4pt}%
  \setlength{\topsep}{0pt}\setlength{\partopsep}{0pt}%
  \begin{quote}\small\noindent\ignorespaces
}{%
  \end{quote}\vspace{4pt}%
  \noindent\rule{\textwidth}{0.4pt}\par\vspace{4pt}%
}

\begin{document}

\begin{center}
  {\LARGE\bfseries Embedded Graph Flows for Categorical Graph Generation\par}
  \vspace{14pt}
  {\large
    Ethan Ma\textsuperscript{1} \quad
    Zihan Wang\textsuperscript{1} \quad
    Chris Siu Yeung Chow\textsuperscript{2} \quad
    Xinguo Feng\textsuperscript{1} \\[3pt]
    Qingqing Li\textsuperscript{3} \quad
    Rui Jiang\textsuperscript{4} \quad
    Naipeng Dong\textsuperscript{1} \quad
    Guangdong Bai\textsuperscript{5}\par}
  \vspace{10pt}
  {\small
    \affil{1}{School of Electrical Engineering and Computer Science, The University of Queensland, Australia}\\
    \affil{2}{Institute for Molecular Bioscience, The University of Queensland, Australia}\\
    \affil{3}{Qunar.com, China} \qquad
    \affil{4}{Chinese Academy of Sciences, China}\\
    \affil{5}{Department of Computer Science, City University of Hong Kong, Hong Kong SAR, China}\par}
\end{center}
\vspace{10pt}

\begin{abstract}
Generating categorical graphs requires choosing node and edge types that form a coherent structure without depending on node order. Many graph generators encode categories as fixed one-hot vectors, which can impose an artificial geometry in which categories are equidistant. We propose Embedded Graph Flows (EGF), a generative model that learns continuous embeddings for node and unordered-edge categories and transports Gaussian noise towards these learnt endpoints using a permutation-equivariant graph transformer. A terminal readout maps the embeddings back to discrete graph categories. Across molecular benchmarks, EGF achieved competitive performance. On QM9, EGF gives the best result on all four reported metrics among the three methods, including a Fr\'echet ChemNet Distance (FCD) of 0.150, compared with 0.717 for the categorical-diffusion baseline DiGress and 0.812 for the bridge-based baseline GruM. When applied to larger molecules in ZINC250k, EGF retains the lowest maximum mean discrepancy (MMD) using the neighbourhood subgraph pairwise distance kernel (NSPDK), indicating close agreement with the local substructures of the reference molecules. Our code is available at \url{https://github.com/Trusted-System-Lab/EGF}.
\end{abstract}

\vspace{6pt}

		\section{Introduction}
		\label{sec:intro}
	
	Categorical graph generation must choose node and edge types jointly because
	connectivity and type compatibility couple local decisions across the graph. The
	model must also accommodate variable graph sizes and preserve permutation
	symmetry: relabelling the nodes cannot change the graph distribution. These
	requirements are critical in molecular generation, where one incompatible
		atom-bond assignment invalidates an otherwise plausible structure.

\begin{figure}[H]
		\centering
		\includegraphics[width=\textwidth]{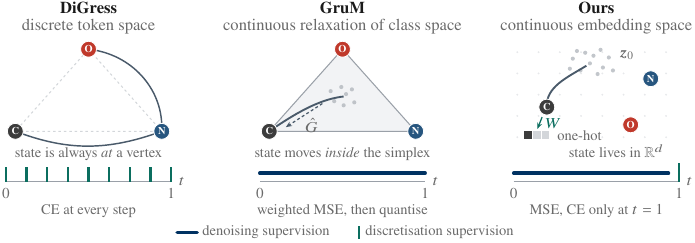}
		\caption{Transport spaces and supervision for DiGress~\cite{vignac2023digress}, GruM~\cite{jo2024grum}, and EGF, illustrated with C, N, and O. DiGress moves between simplex vertices with stepwise cross-entropy. GruM follows an Ornstein-Uhlenbeck bridge mixture over Gaussian-initialised, category-coded arrays and predicts \(\widehat G\) with weighted mean-squared error before quantisation. EGF moves a Gaussian source through \(\mathbb R^d\), using mean-squared error along the path and cross-entropy at \(t=1\). In its panel, \(W\) abbreviates \(W_r\) followed by \(\arg\max\).}
		\label{fig:compare}
	\end{figure}

		Existing methods evolve either categorical node and edge states, continuous arrays tied to
		fixed category codes, or probabilities over fixed one-hot
		endpoints~\cite{vignac2023digress,qin2025defog,eijkelboom2024catflow,jo2024grum}.
		For continuous graph representations, the category encoding fixes the geometry of
		the state space. With one-hot endpoints, all distinct labels remain equidistant
		throughout training. Embedded Language Flows uses a trainable continuous embedding
		space for sequence representations~\cite{hu2026elf}.
		Figure~\ref{fig:compare} illustrates this representational difference. We ask
		whether endpoint geometry can likewise be learned for categorical graphs while
		accommodating padding and preserving undirected-edge and node-relabelling
		symmetries.
 
		\noindent\textbf{Our work.} Embedded Graph Flows (EGF) places node and unordered-edge categories at trainable
		anchors in \(\mathbb R^d\). The anchors define the clean endpoints of the transport
		paths and provide the regression targets during training. A permutation-equivariant
		graph transformer predicts clean embeddings at points sampled along straight conditional paths from
	Gaussian noise. During generation, these predictions define a vector field that is
	integrated as an ordinary differential equation, and a learned terminal
	readout assigns categories.
	For padded, undirected graphs, the mask excludes padded slots from supervision and
	decoding. Each unordered node pair follows one trajectory to one terminal decision,
	and transport remains equivariant under node relabelling.

		\noindent\textbf{Contributions.} EGF makes endpoint geometry trainable by learning node and edge category
	anchors jointly with continuous transport and terminal decoding. Its masked,
	undirected formulation is permutation-equivariant and assigns one shared state to
	each unordered node pair throughout training and generation.
	Empirically, we evaluate EGF on two
	widely used molecular benchmarks~\cite{ramakrishnan2014qm9,sterling2015zinc}: QM9 for small organic molecules and ZINC250k for
	larger drug-like molecules. Under a common evaluator, EGF gives the best QM9 result
	on all four reported metrics. Its Fr\'echet ChemNet Distance (FCD) is \(0.150\),
	compared with \(0.717\) for DiGress and \(0.812\) for GruM, and its maximum mean
	discrepancy (MMD) computed with the neighbourhood subgraph pairwise distance kernel
	(NSPDK) is the lowest of the three on both benchmarks.

	\section{Related Work}
	\label{sec:related}
	
	\subsection{Graph Diffusion and Flow Models}
	
	Flow matching learns a vector field along chosen probability paths without
	simulating complete paths during training~\cite{lipman2023fm}. Graph generators
	instantiate noising and transport in continuous or categorical state spaces. GDSS
	and GruM use real-valued node and edge arrays, through score-based diffusion and an
	Ornstein-Uhlenbeck bridge mixture, respectively~\cite{jo2022gdss,jo2024grum}.
	DiGress applies categorical diffusion~\cite{vignac2023digress}. Discrete flow
	models use continuous-time category transitions~\cite{campbell2024dfm}; DeFoG
	makes the transition rates permutation-equivariant and supports flexible
	sampling~\cite{qin2025defog}. Related molecular
	flow work studies probability-vector, Gaussian-to-one-hot, discrete,
	optimal-transport, and endpoint-consistent
	paths~\cite{dunn2024flowmol,dunn2024discrete,hou2024ggflow,roos2026cfm}.

	EGF transports Gaussian vectors towards learned node and edge anchors with a
	deterministic ordinary
	differential equation, then applies a categorical readout at the endpoint. Learning
	the anchors lets the data shape the relative category geometry, while
	clean-embedding prediction gives transport and readout a shared representation.

	\subsection{Categorical Endpoint Flows}

	CatFlow predicts a categorical distribution over fixed one-hot endpoints; its
	probability-weighted endpoint lies in the simplex and directly defines the
	field~\cite{eijkelboom2024catflow}. In contrast, EGF predicts a clean vector in a learned anchor
	space, derives the continuous field from that prediction, and performs categorical
	readout at \(t=1\).
 
	\subsection{Embedding-Space Generation and Terminal Decoding}
	
	Embedded Language Flows uses straight-line transport in a continuous embedding
	space, predicts clean embeddings during denoising, and trains a terminal categorical
	readout on clean embeddings mixed with Gaussian
	noise~\cite{hu2026elf}. The same network handles denoising and terminal readout.
	EGF adapts this shared continuous-to-discrete design to the node and edge variables
	of a graph.

	Applying this design to graphs requires handling padding, undirected edges, and node
	relabelling. EGF first masks padded coordinates, shares one state between \((i,j)\) and
	\((j,i)\), and makes both transport and readout equivariant to node relabelling. Unlike approaches such as DGAE, which factor generation through a discrete graph auto-encoder and an
	autoregressive model of its codes~\cite{boget2024dgae}, EGF transports continuous
	graph states directly and assigns all active node and unordered-edge categories in
	parallel at the endpoint.    
	
	\section{Method}
	\label{sec:method}
	
	EGF embeds the category at each active node and unordered pair as a continuous
	vector. During training, the graph transformer predicts this clean embedded graph
	from noisy states along the generation path and learns to recover discrete
	categories from clean embeddings mixed with Gaussian noise.

\begin{figure}[H]
			\centering
			\includegraphics[width=\textwidth]{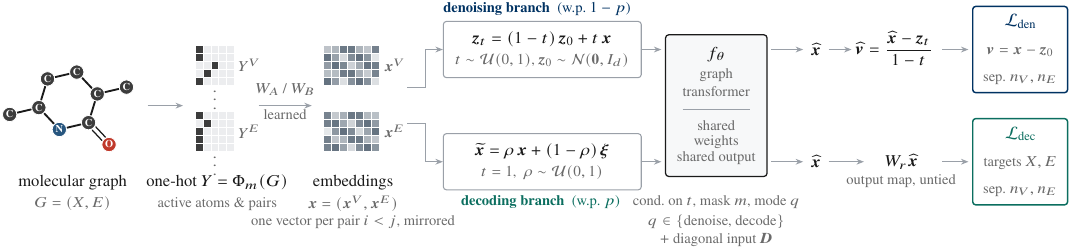}
			\caption{EGF trains a shared graph transformer in denoising and decoding modes. The maps \(W_A,W_B\) embed node and edge categories. The denoising branch predicts clean embeddings along the Gaussian path; the decoding branch feeds noise-perturbed clean embeddings to the same transformer and maps its output to categories with \(W_r\). Unordered-edge states are mirrored, padding is masked, and \(\bm D\) occupies the diagonal input.}
			\label{fig:training}
		\end{figure}

	To generate a graph, EGF starts from Gaussian noise and repeatedly updates its
	continuous node and edge states using the transformer's current clean-embedding
	prediction. A final decoding pass selects the categories. The mask excludes
		padding throughout, and each undirected edge shares one state and one decision
		across its two stored directions. Domain-specific reconstruction and validation
		follow categorical decoding. The two training branches are shown in
		Figure~\ref{fig:training}.

		\subsection{Graph State and Embedding Map}
	\label{subsec:state}
	
	The first step converts a padded categorical graph into the continuous state
	transported by EGF.
	A padded representation reserves \(N=n_{\max}\) node slots for every graph,
	where \(n_{\max}\) is the maximum padded capacity. For any positive integer
	\(k\), write \([k]:=\{1,\ldots,k\}\), so \(i,j\in[N]\) index slots. Let
	\(C_V\) and \(C_E\) denote the numbers of node and edge categories. Write the
	undirected typed graph as \(G=(X,E)\), where \(X\) is the node-category array and
	\(E\) is the symmetric edge-category array.
	Its padded representation is paired with a mask \(m\in\{0,1\}^{N}\), where
	\(m_i=1\) marks an active node slot and \(m_i=0\) marks padding. For \(m_i=1\),
	\(X_i\in[C_V]\) is the node category; for \(i<j\) with \(m_i m_j=1\),
	\(E_{ij}=E_{ji}\in[C_E]\) is the category of the unordered pair between slots
	\(i\) and \(j\). Entries outside these active coordinates are padding and carry
	no graph category.

	To handle node and edge variables uniformly, each active node or unordered pair is
	treated as one categorical coordinate.
	The labels \(V\) and \(E\) in subscripts, superscripts, and values of \(r\)
	distinguish node and edge coordinates. We write
	\(\mathcal A_V(m)\) for active node coordinates and \(\mathcal A_E(m)\)
	for active unordered-pair coordinates, and take their disjoint union as the full
	target set:
	\begin{align*}
		\mathcal A_V(m)
		&=\{i\in[N]:m_i=1\},\\
		\mathcal A_E(m)
		&=\bigl\{\{i,j\}:1\le i<j\le N,\;m_i m_j=1\bigr\},\\
		\mathcal A(m)
		&=\mathcal A_V(m)\ \dot\cup\ \mathcal A_E(m).
	\end{align*}
	Thus a coordinate \(a\in\mathcal A(m)\) is either an active node or an active
	unordered pair and receives a category target during training. Its type is
	\(r(a)\in\{V,E\}\), with \(C_{r(a)}\) possible categories.
	
	The mask defines the active slots and target set \(\mathcal A(m)\); training and
	decoding operate on these coordinates. The edge class \textsc{none} is a valid
	target. The node output alphabet also contains \textsc{absent} for terminal readout;
	the reported training targets exclude this class. The domain adapter interprets both
	classes after terminal decoding.

	For a graph in the training data, EGF records the category at each active
	coordinate and then embeds it for transport.

	\noindent\textbf{One-hot encoding.} For a node coordinate, set \(g_i:=X_i\); for an
	edge coordinate, set \(g_{\{i,j\}}:=E_{ij}\). Thus
	\(g_a\in[C_{r(a)}]\) is the categorical label at coordinate \(a\). We identify
	semantic category names such as \textsc{absent} and \textsc{none} with their fixed
	indices in the corresponding category set. For \(r\in\{V,E\}\) and
	\(c\in[C_r]\), let \(\bm e_c^{(r)}\in\mathbb R^{C_r}\) be the \(c\)-th standard
	basis vector. Let \(\Phi_m\) be the fixed map that collects these vectors over the
	active coordinates:
	\begin{equation}
		Y_a=\bm e_{g_a}^{(r(a))}\in\mathbb R^{C_{r(a)}},
		\qquad
		Y=\Phi_m(G).
		\label{eq:onehot}
	\end{equation}
	Write \(Y=(Y^V,Y^E)\) for the node and edge parts of this
	collection. The fixed map \(\Phi_m\) applies independently at each active
	coordinate. Within each coordinate type, all distinct basis vectors are
	equidistant, so \(\Phi_m\) records category identity with a fixed geometry.
	
	\noindent\textbf{Learnable embedding.} EGF maps these one-hot codes into a shared
	continuous space for transport. Let
	\(W_A\in\mathbb R^{d\times C_V}\) and
	\(W_B\in\mathbb R^{d\times C_E}\) be the learned maps for node (atom in the
	molecular instantiation) and edge (bond) categories, respectively:
	\begin{equation}
		\bm x_a=
		\begin{cases}
			W_A Y_a, & a\in\mathcal A_V(m),\\
			W_B Y_a, & a\in\mathcal A_E(m),
		\end{cases}
		\qquad
		\bm x=(\bm x_a)_{a\in\mathcal A(m)}=(\bm x^V,\bm x^E).
		\label{eq:embed}
	\end{equation}
	The node and edge embedding maps have separate weights and are trained jointly
	with the network. Their columns define the category anchors, whose arrangement in
	\(\mathbb R^d\) is determined during training.

	During generation, the sampler draws \(K\le N\) from the empirical training-size
	distribution and constructs a mask \(m\) with
	\(K=|\mathcal A_V(m)|=\sum_{i=1}^{N}m_i\) active slots.

	To distinguish categorical codes from continuous states, bold symbols denote
	continuous vectors or arrays of vectors. We use \(Y\) for the deterministic
	one-hot encoding, \(\bm Z\) for a random graph state,
	\(\bm z\) for a realisation, and \(\bm D\) for a node-aligned state held on the
	edge-array diagonal. Category
	labels, mask entries, indices, and times are scalar.

	Active-coordinate notation is convenient for graph semantics and losses, whereas
	the transformer receives fixed-size arrays. For
	computation, a graph state is stored as a zero-padded node array and
	a symmetric zero-padded edge array. Its state space is
	\begin{equation}
		\mathcal Z_m:=\left\{
		\bm z=(\bm z^V,\bm z^E)\in\mathbb R^{N\times d}\times
		\mathbb R^{N\times N\times d}:
		\begin{array}{l}
		\bm z_i^V=\bm 0\ \text{if }m_i=0,\\
		\bm z_{ij}^E=\bm z_{ji}^E,\\
		\bm z_{ij}^E=\bm 0\ \text{if }i=j\ \text{or }m_i m_j=0
		\end{array}
		\right\}.
		\label{eq:state-space}
	\end{equation}
	Here \(\bm 0\in\mathbb R^d\) is the zero vector. For
	\(a\in\mathcal A(m)\), the notation \(\bm z_a\) denotes its unique active vector;
	an unordered-pair vector is stored at both \((i,j)\) and \((j,i)\) but counted
	once. We place \(\bm x\) from \eqref{eq:embed} in this dense representation by
	mirroring active pair vectors and setting all other entries to zero. Thus
	\(\bm x\in\mathcal Z_m\) is the clean endpoint of the flow defined in
	Section~\ref{subsec:path}.
	
	\noindent\textbf{Diagonal state.} The dense edge array reserves each active diagonal
	slot for an additional state aligned with node \(i\). Let
	\(\bm D_i\in\mathbb R^d\) denote its diagonal vector and let
	\(\bm D=(\bm D_i)_{i\in\mathcal A_V(m)}\) collect these vectors.
	For \(\bm z\in\mathcal Z_m\), write
	\(\overline{\bm z}=(\bm z,\bm D)\) for the complete network input, understood
	as the node array \(\bm z^V\) and the dense edge array obtained by inserting
	\(\bm D_i\) at each active diagonal position of \(\bm z^E\). Write
	\(\bm x_{\textsc{none}}:=W_B
	\bm e_{\textsc{none}}^{(E)}\) for the embedded no-edge vector used as the endpoint
	of this channel. The diagonal state depends on the stage of the workflow. For every
	\(i\in\mathcal A_V(m)\), training uses
	\(\bm D_{t,i}^{\mathrm{train}}=t\bm x_{\textsc{none}}\), sampling starts from
	\(\bm D_{0,i}^{\mathrm{sample}}=\bm 0\), and decoding uses
	\(\bm D_i^{\mathrm{dec}}=\bm x_{\textsc{none}}\). The corresponding collections
	are denoted \(\bm D_t^{\mathrm{train}}(m)\),
	\(\bm D_0^{\mathrm{sample}}(m)\), and \(\bm D^{\mathrm{dec}}(m)\); a subscript
	\(b\) means that mask \(m_b\) is used.
	During sampling, this state evolves according to the update in
	Section~\ref{sec:sampler}. The network can use this channel in attention biases and
	incident-edge summaries. Training targets and graph reconstruction use the node and
	off-diagonal edge coordinates.
	
	\subsection{Source and Conditional Path}
	\label{subsec:path}
	
	The embedding map supplies the clean endpoint \(\bm x\). EGF uses a standard
	Gaussian source in the learned embedding space. Conditional on \(m\), let
	\(\bm Z_1\in\mathcal Z_m\) be the random clean endpoint induced by the graph
	distribution, and let \(\bm x\) be one realisation. The source is drawn
	independently of \(\bm Z_1\) given \(m\), and the two are connected by the straight
	conditional path of rectified flow:
	\begin{equation}
		\bm Z_{0,a}\overset{\mathrm{iid}}{\sim}\mathcal N(\bm 0,I_d)
		\ \ \text{for } a\in\mathcal A(m),
		\qquad
		\bm Z_t=(1-t)\bm Z_0+t\bm Z_1,
		\ \ t\in[0,1].
		\label{eq:path}
	\end{equation}
	Here \(I_d\) is the \(d\times d\) identity matrix. For an unordered edge
	coordinate \(a=\{i,j\}\), the single draw \(\bm Z_{0,a}\) is stored symmetrically as
	\(\bm Z^E_{0,ij}=\bm Z^E_{0,ji}:=\bm Z_{0,a}\). Both stored directions of each
	active pair therefore share one edge-noise vector. Conditioned on
	\(\bm Z_1=\bm x\), \eqref{eq:path} interpolates linearly from \(\bm Z_0\) to
	\(\bm x\). The path remains in the masked state space
	\eqref{eq:state-space} throughout transport, and categorical projection occurs at
	the endpoint.
	
	\subsection{Embedding-Space Flow Field}
	
	At generation time, the clean endpoint is unknown, so the sampler needs a field
	determined by its current state. For a fixed training endpoint \(\bm x\),
	differentiating \eqref{eq:path} gives the conditional field. Averaging over the
	endpoints compatible with the intermediate state gives the marginal field used for
	generation. Both fields belong to \(\mathcal Z_m\):
	\begin{equation}
		\begin{aligned}
		\bm u_t(\bm z\mid \bm Z_1=\bm x,m)
		&=\frac{\bm x-\bm z}{1-t},\\
		\bm\mu_t(\bm z,m)
		&:=\mathbb E[\bm Z_1\mid \bm Z_t=\bm z,t,m],\\
		\bm u_t(\bm z\mid m)
		&=\frac{\bm\mu_t(\bm z,m)-\bm z}{1-t},
		\qquad 0\le t<1.
		\end{aligned}
		\label{eq:marginal}
	\end{equation}
	Here \(\bm\mu_t(\bm z,m)\in\mathcal Z_m\) is the posterior mean clean endpoint.
	The marginal field points from the current state toward this mean with the
	\(1/(1-t)\) rectified-flow scaling.
	
	\noindent\textbf{Model output and field.} The posterior mean in
	\eqref{eq:marginal} is not available directly, so EGF approximates it with the graph
	transformer. Let \(\theta\) denote the trainable network parameters. The network
	receives a mode indicator
	\(q\in\{\mathrm{denoise},\mathrm{decode}\}\), matching Figure~\ref{fig:training}.
	The first value selects interior denoising and the second endpoint decoding. Write
	\(f_\theta=(f_\theta^V,f_\theta^E)\) for its full dense node and edge output
	streams in \(\mathbb R^{N\times d}\) and
	\(\mathbb R^{N\times N\times d}\), respectively. Restricting these streams to
	\(\mathcal A(m)\) gives the shared output
	\(\widehat{\bm x}_\theta(\overline{\bm z},t,m,q)\), whose component at coordinate
	\(a\) is \(\widehat{\bm x}_{\theta,a}\in\mathbb R^d\). In denoising mode, EGF
	interprets the output as a clean-endpoint estimate and converts it into
	the model field:
	\begin{equation}
		\widehat{\bm v}_\theta(\overline{\bm z},t,m)
		=\frac{\widehat{\bm x}_\theta(
		\overline{\bm z},t,m,\mathrm{denoise})-\bm z}{1-t},
		\qquad 0\le t<1 .
		\label{eq:field}
	\end{equation}
	By predicting the clean endpoint, the network supplies the same \(d\)-dimensional
	representation to both branches; the decoding map converts it to category scores.
	Prior work reports poor shared-weight categorical readout under direct velocity
	parameterisation~\cite{hu2026elf}.
	
	Under squared-error denoising, this clean-endpoint parameterisation recovers the
	field above. Specifically, for fixed embedding maps and a complete training input
	\(\overline{\bm z}=(\bm z,\bm D_t^{\mathrm{train}}(m))\), the population minimiser of the
	denoising mean-squared error below recovers the posterior mean on the active
	coordinates:
	\[
		\begin{aligned}
		\widehat{\bm x}^{\,*}(\overline{\bm z},t,m)
		&=\bm\mu_t(\bm z,m),\\
		\frac{\widehat{\bm x}^{\,*}(\overline{\bm z},t,m)-\bm z}{1-t}
		&=\bm u_t(\bm z\mid m).
		\end{aligned}
	\]
	Equation~\eqref{eq:field} therefore recovers the marginal field in
	\eqref{eq:marginal} from the clean-endpoint estimate. During training,
	\(\bm D_t^{\mathrm{train}}(m)\) is a deterministic function of \(t\) and \(m\); the
	complete input carries the same conditioning information as \(\bm Z_t,t,m\).
	
	\subsection{Two-Branch Training Objective}
	\label{subsec:objective}
	
	Denoising learns transport through the embedding space, while decoding learns the
	terminal categorical readout. Both branches share \(f_\theta\), and a mode is
	selected separately for each graph. Let
	\(\{(G_b,m_b):G_b=(X_b,E_b)\}_{b=1}^{B}\) be a minibatch of \(B\) padded graphs, let
	\(m_b=(m_{b1},\ldots,m_{bN})\) be graph \(b\)'s mask, and let \(g_{b,a}\) be its
	target category at coordinate \(a\).

	Node and edge errors are normalised separately. For \(r\in\{V,E\}\), define
	\(n_r:=\sum_{b=1}^{B}|\mathcal A_r(m_b)|\), the number of supervised
	coordinates of type \(r\) in the full minibatch, and set
	\(\lambda_V=\lambda_E=1\). Here \(n_V\) counts active nodes and \(n_E\) counts
	active unordered pairs. The number of pairs grows quadratically with graph size,
	while the number of nodes grows linearly; the separate counts prevent this growth
	from reweighting the two coordinate types. Let
	\(p\in[0,1]\) be the probability that a graph is assigned to endpoint decoding.
	
	\noindent\textbf{Denoising branch (probability \(1-p\)).} Draw
	\(t_b\sim\mathcal U(0,1)\) independently for each graph, where
	\(\mathcal U(0,1)\) is the uniform distribution on \((0,1)\). Draw a source realisation \(\bm z_{0,b}\),
	form \(\bm z_{t_b,b}=(1-t_b)\bm z_{0,b}+t_b\bm x_b\) by \eqref{eq:path}, and
	append \(\bm D_{t_b}^{\mathrm{train}}(m_b)\) to obtain
	\(\overline{\bm z}_{t_b,b}\). Write
	\(\widehat{\bm v}_{\theta,a}\) for the \(d\)-dimensional component of \eqref{eq:field}
	at coordinate \(a\). Since the target velocity along \eqref{eq:path} is
	\(\bm x_{b,a}-\bm z_{0,b,a}\), the denoising loss is
	\begin{equation}
		\mathcal L_{\mathrm{den}}
		=\sum_{r\in\{V,E\}}\frac{\lambda_r}{n_r}
		 \sum_{b=1}^{B}\sum_{a\in\mathcal A_r(m_b)}
		 \left\lVert
		 \widehat{\bm v}_{\theta,a}(\overline{\bm z}_{t_b,b},t_b,m_b)
		 -\bigl(\bm x_{b,a}-\bm z_{0,b,a}\bigr)
		 \right\rVert_2^{2}.
		\label{eq:lden}
	\end{equation}
	Here \(\lVert\cdot\rVert_2\) is the Euclidean norm. Under \eqref{eq:field}, the
	velocity loss is equivalent to clean-embedding regression weighted by
	\((1-t_b)^{-2}\). This weight grows as \(t_b\) approaches \(1\); the open uniform
	distribution excludes the singular endpoint itself.
	
	\noindent\textbf{Decoding branch (probability \(p\)).} This branch trains the
	categorical readout at \(t=1\). For every graph \(b\), draw one
	\(\rho_b\sim\mathcal U(0,1)\). For each active coordinate \(a\), draw
	\(\bm\xi_{b,a}\sim\mathcal N(\bm 0,I_d)\), then mix the clean embedding with
	Gaussian noise:
	\[
		\widetilde{\bm x}_{b,a}
		=\rho_b\bm x_{b,a}+(1-\rho_b)\bm\xi_{b,a}.
	\]
	For an edge, one noise vector is drawn per unordered pair and mirrored. Using the
	fixed decoding value for \(\bm D\) defined in
	Section~\ref{subsec:state}, write
	\(\overline{\widetilde{\bm x}}_b
	:=(\widetilde{\bm x}_b,\bm D^{\mathrm{dec}}(m_b))\) for the complete input.

	For \(r\in\{V,E\}\), the learned output map
	\(W_r\in\mathbb R^{C_r\times d}\) converts a \(d\)-dimensional output into
	\(C_r\) unnormalised category scores. The output maps \(W_V,W_E\) are untied from
	the embedding maps \(W_A,W_B\). Applying softmax gives a probability
	distribution over the \(C_r\) categories. For each active coordinate, let
	\(\bm\ell_{\theta,a}\in\mathbb R^{C_{r(a)}}\) denote the logits and let
	\(\bm\pi_{\theta,a}\in[0,1]^{C_{r(a)}}\) denote the resulting probabilities,
	\[
		\bm\ell_{\theta,a}(\overline{\bm z},m)
		:=W_{r(a)}
		\widehat{\bm x}_{\theta,a}(
		\overline{\bm z},1,m,\mathrm{decode}),
		\qquad
		\bm\pi_{\theta,a}(\overline{\bm z},m)
		:=\operatorname{softmax}\!\bigl(\bm\ell_{\theta,a}(\overline{\bm z},m)\bigr).
	\]
	The decoding cross-entropy is
	\begin{equation}
		\mathcal L_{\mathrm{dec}}
		=-\sum_{r\in\{V,E\}}\frac{\lambda_r}{n_r}
		 \sum_{b=1}^{B}\sum_{a\in\mathcal A_r(m_b)}
		 \log\!\left[
		 \bm\pi_{\theta,a}(\overline{\widetilde{\bm x}}_b,m_b)
		 \right]_{g_{b,a}}.
		\label{eq:ldec}
	\end{equation}
	The two branches use the same coordinate counts and type weights. Padded
	coordinates and diagonal entries remain outside both losses.
	
	Taking the expectation over branch assignments gives the minibatch objective
	\begin{equation}
		\mathcal L
		=(1-p)\,\mathcal L_{\mathrm{den}}
		+p\,\mathcal L_{\mathrm{dec}} .
		\label{eq:ltotal}
	\end{equation}
	During training, each graph samples a mode \(q_b\) with denoising and decoding
	probabilities \(1-p\) and \(p\), then contributes the corresponding branch loss. The
	resulting minibatch loss is a Monte Carlo estimate of \eqref{eq:ltotal}. Both modes
	are processed in one batched forward and backward pass per step. The optimiser
	updates \(\theta\), \(W_A,W_B\), and \(W_V,W_E\) in the same step.
	
		\subsection{Symmetry-Aware Graph Network}
	
	Because a graph has no privileged node ordering, the network must transform
	equivariantly when padded slots are relabelled. It therefore applies shared
	transformations across slots and receives no absolute slot identifiers. Diagonal
	entries of the dense stream \(f_\theta^E\) are diagonal outputs that lie outside
	\(\mathcal A(m)\). Let \(P\) be
	any \(N\times N\) permutation matrix that reorders the padded slots, and let
	superscript \(\top\) denote matrix transpose. Multiplication by \(P\) and
	\(P^\top\) acts on the two slot-index axes of a dense edge array and leaves its
	embedding axis unchanged. Define
	\(P\cdot\overline{\bm z}:=(P\bm z^V,P\overline{\bm z}^EP^\top)\). The
	architecture satisfies
	\[
		f_\theta(P\cdot\overline{\bm z},t,Pm,q)
		=\left(Pf_\theta^V(\overline{\bm z},t,m,q),
		        Pf_\theta^E(\overline{\bm z},t,m,q)P^\top\right),
	\]
	for either mode. Reordering the slots and mask only reorders the corresponding
	node and edge outputs.

	Every layer is modulated by a numerical encoding of time and a learned embedding
	of the current mode. Each layer updates the edge stream using attention and the
	features of each node pair. It then averages the two stored directions to restore
	symmetry. The final edge output is symmetrised once more.
	
	Current edge features also affect node attention through an additive score term.
	For any attention head \(h\) of width \(d_h\), let
	\(\bm Q_i^{(h)},\bm K_j^{(h)}\in\mathbb R^{d_h}\) be the query and key vectors
	for nodes \(i\) and \(j\), and let \(\gamma_{ij}^{(h)}\in\mathbb R\) be the
	learned scalar contribution from their current edge features. The pre-mask
	attention score is
	\[
		s^{(h)}_{ij}
		=\frac{\langle\bm Q_i^{(h)},\bm K_j^{(h)}\rangle}{\sqrt{d_h}}
		+\gamma^{(h)}_{ij}.
	\]
	Here \(\langle\cdot,\cdot\rangle\) is the vector dot product. Query-key
	normalisation applies separate layer normalisations to \(\bm Q_i^{(h)}\) and
	\(\bm K_j^{(h)}\) before their dot product. A nonnegative hyperparameter
	\(c_{\mathrm{cap}}\) optionally caps the attention scores. When
	\(c_{\mathrm{cap}}>0\), the score is replaced by
	\(c_{\mathrm{cap}}\tanh(s^{(h)}_{ij}/c_{\mathrm{cap}})\) before padded columns
	are masked; \(c_{\mathrm{cap}}=0\) leaves the score uncapped.
	
	Permutation equivariance handles node relabelling; exact undirected symmetry also
	requires the two stored directions of an edge to agree. Each active pair \(i<j\)
	follows one shared trajectory from its source draw to its terminal decision. After
	every transport update, the sampler averages its two stored directions to remove
	numerical asymmetry. The diagonal state follows the separate convention above.
	
	A domain adapter may provide additional node features. These features reorder with
	the node slots and are used only as network inputs.
	
	\subsection{Sampling and Terminal Decoding}
	\label{sec:sampler}
	
	During generation, denoising predictions drive the numerical trajectory, followed
	by one decoding pass at the endpoint.
	Choose a positive integer \(S\) and a sequence of times
	\(0=t_0<t_1<\cdots<t_S=1\). These times divide the path into \(S\)
	numerical integration intervals, indexed by \(s=0,\ldots,S-1\). Let
	\(\bm z_s^{\mathrm{sample}}\in\mathcal Z_m\) denote the realised sampling state
	at time \(t_s\). Unlike the training state \(\bm Z_t\) in \eqref{eq:path}, it is
	generated without a known clean endpoint. Initialise
	\(\bm z_0^{\mathrm{sample}}\) with a realisation of the Gaussian source and use
	the zero diagonal state \(\bm D_0^{\mathrm{sample}}(m)\) from
	Section~\ref{subsec:state}.

	At interval \(s\), one network evaluation on
	\(\overline{\bm z}_s^{\mathrm{sample}}
	=(\bm z_s^{\mathrm{sample}},\bm D_s^{\mathrm{sample}})\) produces the clean-endpoint
	estimate on active coordinates and the diagonal output at every
	active node,
	\[
		\widehat{\bm x}_s
		:=\widehat{\bm x}_\theta(
		\overline{\bm z}_s^{\mathrm{sample}},t_s,m,\mathrm{denoise}),
		\qquad
		\bm o^D_{s,i}
		:=\bigl[f_\theta^E(
		\overline{\bm z}_s^{\mathrm{sample}},t_s,m,\mathrm{denoise})\bigr]_{ii},
		\quad i\in\mathcal A_V(m).
	\]
	These outputs drive the explicit Euler updates
	\begin{align}
		\bm z_{s+1}^{\mathrm{sample}}
		&=\bm z_s^{\mathrm{sample}}+(t_{s+1}-t_s)\,
		\frac{\widehat{\bm x}_s-\bm z_s^{\mathrm{sample}}}
		{\max\{1-t_s,10^{-4}\}},
		\label{eq:euler}\\[4pt]
		\bm D_{s+1,i}^{\mathrm{sample}}
		&=\bm D_{s,i}^{\mathrm{sample}}+(t_{s+1}-t_s)\,
		\frac{\bm o^D_{s,i}-\bm D_{s,i}^{\mathrm{sample}}}
		{\max\{1-t_s,10^{-4}\}}.
		\label{eq:diag-euler}
	\end{align}
	Let \(\bm o_s^D:=(\bm o^D_{s,i})_{i\in\mathcal A_V(m)}\) collect the diagonal
	outputs. Both updates use the same time increment and denominator safeguard.
	Equation~\eqref{eq:euler} updates the node and off-diagonal edge state;
	\eqref{eq:diag-euler} evolves the diagonal state. After each update, the sampler
	averages the two stored directions of every
	off-diagonal edge, zeros every node entry outside \(\mathcal A_V(m)\) and
	every off-diagonal edge entry whose unordered pair lies outside
	\(\mathcal A_E(m)\), and retains diagonal vectors only for slots in
	\(\mathcal A_V(m)\). After source initialisation, the Euler rollout is
	deterministic.
	
	Every grid specified in Section~\ref{sec:sampler-settings} satisfies
	\(1-t_{S-1}>10^{-4}\). Setting \(t_S=1\) in \eqref{eq:euler} therefore gives
	\begin{equation}
		\bm z_S^{\mathrm{sample}}=\widehat{\bm x}_{S-1},
		\label{eq:zs}
	\end{equation}
	so the last transport update places every active coordinate at the clean-embedding
	estimate from the evaluation at \(t_{S-1}\). Before terminal decoding, the sampler
	replaces the diagonal state with the fixed value used in decoding training and
	forms \(\overline{\bm z}_S^{\mathrm{dec}}
	:=(\bm z_S^{\mathrm{sample}},\bm D^{\mathrm{dec}}(m))\). One further network evaluation
	in decoding mode at \(t=1\) produces a vector for
	each active coordinate. For a coordinate of type \(r\), multiplication by
	\(W_r\) gives one score per category, and \(\arg\max\) selects the
	category with the largest score. The operation is applied once per active node
	and once per active pair \(i<j\); each edge decision is then copied to \((j,i)\).
	
\begin{figure}[H]
		\centering
		\includegraphics[width=\textwidth]{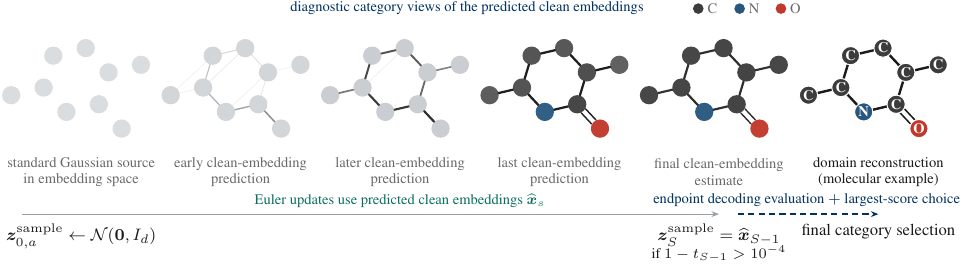}
		\caption{EGF sampling uses \(S\) denoising calls from a Gaussian source followed by one terminal decoding call. Molecular reconstruction and sanitisation follow category selection; the intermediate panels visualise predicted clean embeddings.}
		\label{fig:trajectory}
	\end{figure}

	The shared network is evaluated in decoding mode at \(t=1\) during training and
	sampling. Training supplies clean embeddings mixed with Gaussian noise; generation
	supplies the numerical-rollout endpoint. A rollout with \(S\) intervals uses \(S\)
	denoising evaluations and one decoding evaluation.
	Figure~\ref{fig:trajectory} illustrates the rollout, and
	Algorithm~\ref{alg:sample} gives the complete procedure.

\begin{algorithm}[H]
		\caption{EGF sampling and terminal decoding.}
		\label{alg:sample}
		\small
		\begin{enumerate}[leftmargin=1.6em,itemsep=2pt,topsep=2pt]
		\item Draw the active-slot count \(K\) from the empirical training-size distribution,
			      construct \(m\), and draw the masked source with one Gaussian vector
			      per active node and per active pair \(i<j\), mirrored to \((j,i)\).
		\item Set every component of \(\bm D_0^{\mathrm{sample}}(m)\) to \(\bm 0\). For
			      \(s=0,\ldots,S-1\), evaluate the network at
			      \((\overline{\bm z}_s^{\mathrm{sample}},t_s,m,\mathrm{denoise})\), form
			      \(\widehat{\bm x}_s\) and the diagonal outputs \(\bm o_s^D\), apply
			      \eqref{eq:euler} and
			      \eqref{eq:diag-euler}, then restore edge symmetry and reapply the
			      masking rule above.
			\item Form
			      \(\overline{\bm z}_S^{\mathrm{dec}}
			      =(\bm z_S^{\mathrm{sample}},\bm D^{\mathrm{dec}}(m))\) and evaluate
			      the same network once more at
			      \((\overline{\bm z}_S^{\mathrm{dec}},1,m,\mathrm{decode})\). Apply
			      \(W_V\) to node outputs and \(W_E\) to edge
			      outputs, select the largest score at each active node and once for each
			      pair \(i<j\), then mirror the edge decisions.
			\item Pass the projected categorical graph to the domain adapter for readout
			      interpretation, reconstruction and validity assessment.
		\end{enumerate}
	\end{algorithm}
	
	\subsection{Categorical Readout and Domain Reconstruction}
	
	The numerical rollout ends with a continuous state, whereas reconstruction operates
	on a categorical graph.

	\noindent\textbf{Terminal categorical projection.} Apply the logit map \(\bm\ell_{\theta,a}\) of
	Section~\ref{subsec:objective} to the terminal input \(\overline{\bm z}_S^{\mathrm{dec}}\). The decoded
	graph and its category at each active coordinate are
	\[
		\widehat G=(\widehat X,\widehat E),
		\qquad
		\widehat g_a
		=\underset{c\in[C_{r(a)}]}{\arg\max}\;
		\bigl[\bm\ell_{\theta,a}(\overline{\bm z}_S^{\mathrm{dec}},m)\bigr]_c.
	\]
	The selected labels \(\widehat g_i\) and \(\widehat g_{\{i,j\}}\) populate
	\(\widehat X_i\) and \(\widehat E_{ij}\), respectively. The decoded graph is paired
	with the mask \(m\). Each active node
	receives one decision, and each active
	pair \(i<j\) receives one decision that is copied to \((j,i)\); padded coordinates
	remain excluded. The result is a typed categorical graph. Domain-specific
	interpretation, reconstruction, and validity assessment follow.
	
	Applying EGF to a new domain requires node and edge alphabets, special readout
	categories, reconstruction rules, and any adapter features.
	
	\noindent\textbf{Molecular readout.} In the reported QM9 and ZINC250k experiments,
	a node labelled \textsc{absent} is removed with its incident edges,
	\textsc{none} denotes no bond, and padding never enters the graph. The remaining
	atom and bond categories define a molecular graph. The molecular graph is first
	sanitised without repair. If sanitisation fails, the optional repair procedure is
	attempted. All reported validity values are measured before repair, and repaired
	results are kept separate.

	The molecular adapter also converts each current dense edge vector to provisional
	category probabilities
	\(\operatorname{softmax}(W_E\overline{\bm z}^E_{ij})\). For each node, it
	sums the non-\textsc{none} probability components over active partner slots,
	computes a bond-order-weighted total, and divides both summaries by the fixed scale
	4. It also computes the mean probability of each edge category over those slots.
	These summaries are used as network input features.

\section{Experiments}
	\label{sec:experiments}

	We evaluate EGF on two molecular graph benchmarks and rescore released DiGress and
	GruM samples through the same pipeline.

	\subsection{Experimental Setup}
	\label{sec:setup}
    
	\subsubsection{Datasets and Encodings}
 
	QM9 is a benchmark of small organic molecules, and ZINC250k is a benchmark of
	larger drug-like molecules~\cite{ramakrishnan2014qm9,sterling2015zinc}. After
	processing, they contain 129{,}012 and 249{,}455 molecules, respectively. Using
	random seed 0, we permute each dataset, assign the first 10{,}000 molecules to
	the test set and the next 10{,}000 to validation, and retain 109{,}012 and
	229{,}455 training examples. The padded capacities match the largest graph in
	each dataset: \(N=n_{\max}=9\) for QM9 and \(N=n_{\max}=38\) for ZINC250k.
	These capacities provide \(\binom{9}{2}=36\) and \(\binom{38}{2}=703\)
	unordered node-pair positions per graph. In the reported QM9 graphs, hydrogen atoms
	are implicit and inferred from the atom and bond types.

	We evaluate aromatic and kekulised graph encodings. The \emph{aromatic} encoding
	has \(C_E=5\) edge categories
	(\textsc{none}, single, double, triple, and aromatic). The \emph{kekulised}
	encoding represents aromatic bonds by alternating single and double bonds and
	therefore has \(C_E=4\). Aromatic node categories combine the element, formal
	charge, and number of aromatic hydrogens; kekulised node categories use the
	element and formal charge. For the kekulised encoding, we kekulise a copy of each
	molecule before encoding; sanitisation later restores aromaticity and
	implicit-hydrogen rules. Neither encoding represents stereochemistry, isotopes,
	or radical-electron counts.
	Excluding \textsc{absent}, the node-category alphabets contain 19 and 10 classes
	for the QM9 aromatic and kekulised encodings, and 28 and 17 for the ZINC250k
	encodings; including that readout class gives \(C_V=20,11,29,18\), respectively.
	Padding remains controlled by the mask. Both encodings retain canonical aromatic
	Simplified Molecular Input Line Entry System (SMILES) strings as their reference
	representation.
 
		\subsubsection{Optimisation and Inference Settings}
	\label{sec:sampler-settings}
 
	Training uses batches of 256 graphs and the AdamW optimiser with learning rate
	\(2\times10^{-4}\), weight decay 0.01, and moment-decay coefficients
	\((\beta_1,\beta_2)=(0.9,0.95)\). We use 2{,}000 learning-rate warm-up steps,
	clip the gradient norm at 1.0, apply no dropout, and use random seed 0. The
	exponential moving average of the network parameters has decay 0.9999 and
	is used for periodic sample generation. The node and edge embedding maps
	\(W_A,W_B\), their independent output maps \(W_V,W_E\), and the graph network are optimised
	together. As defined in Section~\ref{subsec:objective}, each graph is assigned to
	the decoding branch with probability \(p\), and both branch types can occur in
	one minibatch. Training uses automatic 16-bit brain floating-point (bfloat16)
	conversion for supported operations; sampling and evaluation do not. Training runs
	on one node with eight NVIDIA B200 GPUs.

		\subsubsection{Evaluation Metrics}
	\label{sec:metrics}
 
	Every value in Section~\ref{sec:comparison} uses the same metric implementations,
	test reference, generated-sample count, and strict validity rule. Validity uses all
	generated graphs as its denominator; FCD, NSPDK-based MMD, and scaffold similarity
	use one fixed pool of valid molecules per dataset.

	\noindent\textbf{Validity.} Strict validity is the percentage of generated graphs that
	reconstruct, without correction, to one connected molecule accepted by the
	chemical sanitisation software; a graph that produces multiple disconnected
	fragments is invalid under this rule. An aromatic repair heuristic is attempted only
	after the initial check fails. All
	tabulated validity values are measured before repair.

	\noindent\textbf{Distributional and scaffold agreement.} FCD
	compares the means and covariances of features produced by a pretrained molecular
	network for generated and reference molecules~\cite{preuer2018fcd}, evaluated
	against the corresponding test reference at a common sample count. The
	NSPDK compares local graph neighbourhoods~\cite{costa2010nspdk}; we use it within
	MMD and report the unbiased finite-sample estimate of squared MMD. Molecules that
	cannot be parsed or converted to the
	required graph features are omitted. Scaffold similarity follows the Molecular
	Sets (MOSES) convention~\cite{polykovskiy2020moses} used by the
	baselines~\cite{jo2022gdss,jo2024grum}: index the union of all Bemis-Murcko
	scaffolds appearing in the reference or generated molecules, each holding a
	molecule's ring systems and the linkers between them~\cite{bemis1996scaffold}. If
	that union contains \(M\) scaffolds and
	\(\bm c^{\mathrm{ref}},\bm c^{\mathrm{gen}}\in\mathbb R_{\ge0}^{M}\) are the
	corresponding count vectors, scaffold similarity is their cosine
	\[
		\mathrm{Scaffold}
		=\frac{\langle\bm c^{\mathrm{ref}},\bm c^{\mathrm{gen}}\rangle}
		      {\lVert\bm c^{\mathrm{ref}}\rVert_2
		       \,\lVert\bm c^{\mathrm{gen}}\rVert_2},
	\]
	which removes the overall count scale and compares relative scaffold frequencies.
 
	\subsection{Comparison with Prior Methods}
	\label{sec:comparison}

	\begin{table}[H]
		\centering
		\caption{QM9 and ZINC250k results using the evaluation settings in Section~\ref{sec:metrics}; training settings are method-specific.}
		\label{tab:comparison}
		\resizebox{\textwidth}{!}{%
		\begin{tabular}{lcccccccc}
		\toprule
		& \multicolumn{4}{c}{QM9 (at most 9 nodes)}
		& \multicolumn{4}{c}{ZINC250k (at most 38 nodes)} \\
		\cmidrule(lr){2-5}\cmidrule(lr){6-9}
		\textbf{Method}
		& \textbf{Valid (\%)\(\uparrow\)}
		& \textbf{FCD\(\downarrow\)}
		& \textbf{NSPDK MMD\(\downarrow\)}
		& \textbf{Scaffold\(\uparrow\)}
		& \textbf{Valid (\%)\(\uparrow\)}
		& \textbf{FCD\(\downarrow\)}
		& \textbf{NSPDK MMD\(\downarrow\)}
		& \textbf{Scaffold\(\uparrow\)} \\
		\midrule
		DiGress \cite{vignac2023digress}
		& 98.92 & 0.717 & 0.0031 & 0.7984
		& 82.79 & 2.803 & 0.0032 & 0.4398 \\
		GruM \cite{jo2024grum}
		& 99.69 & 0.812 & 0.0036 & 0.8086
		& \textbf{98.65} & \textbf{2.276} & 0.0028 & \textbf{0.5802} \\
		\midrule
		\textbf{EGF (ours)}
		& \textbf{99.87} & \textbf{0.150} & \textbf{0.0019} & \textbf{0.9936}
		& 96.56 & 3.326 & \textbf{0.0024} & 0.5546 \\
		\bottomrule
		\end{tabular}}
	\end{table}
 
	\noindent\textbf{QM9.} EGF has the best value on all four metrics. Its validity is
	99.87\%, compared with 98.92\% for DiGress and 99.69\%
	for GruM. FCD is 0.150, compared with 0.717 and 0.812; NSPDK-based MMD is 0.0019,
	compared with 0.0031 and 0.0036; and scaffold similarity is 0.9936, compared with
	0.7984 and 0.8086.
 
	\noindent\textbf{ZINC250k.} EGF has FCD 3.326, compared
	with 2.276 for GruM and 2.803 for DiGress, and NSPDK-based MMD 0.0024, compared
	with 0.0028 and 0.0032. Its validity is 96.56\%, above DiGress but below GruM.
	Among the three rows, its scaffold similarity of 0.5546 is above DiGress at 0.4398
	and below GruM at 0.5802.
 
	\section{Discussion}
	\label{sec:discussion}
 
	The results indicate that the molecular metrics capture complementary aspects of the generated graph distribution. NSPDK-based MMD compares local atom-bond graph neighbourhoods and their relative distances~\cite{costa2010nspdk}, whereas scaffold similarity measures agreement in the frequencies of Bemis--Murcko ring--linker frameworks~\cite{bemis1996scaffold,polykovskiy2020moses}.
	FCD instead compares the means and covariances of learned ChemNet
	features~\cite{preuer2018fcd}. The strong QM9 results therefore suggest that EGF captures both local graph structure and broader molecular distributional properties on small molecular graphs. In ZINC250k, the maximum graph size increases from 9 nodes and 36 unordered pairs in QM9 to 38 nodes and 703 unordered pairs in ZINC250k. EGF maintains strong agreement on graph-structural metrics, such as NSPDK-based MMD, while its higher FCD and lower validity than GruM suggest that matching substructures does not ensure broader distributional agreement or chemical validity. From a molecular perspective, larger graphs introduce more atom--bond assignments that must jointly satisfy valence, connectivity, and aromaticity constraints. From a graph-generation perspective, the same result may reflect the increasing difficulty of jointly modelling node and edge categories as graph size and categorical complexity grow.

\section{Limitations and Future Work}
	\label{sec:limits}
	
	A component study is needed to separate the effects of learned endpoint geometry,
	the graph transformer, two-branch training, and terminal readout. The embedding
	stage is lightweight: atom and bond categories are mapped to endpoint anchors by
	one linear transformation each, \(W_A\) and \(W_B\), so every occurrence of a
	category shares the same endpoint regardless of its local chemical environment. The
	graph transformer then provides contextual processing over these anchors. This shared
	endpoint geometry may be less expressive across the broader range of environments found in
	larger molecules and may contribute to EGF's higher ZINC250k FCD. Future work could
	replace the category-only endpoint embeddings with richer, context-dependent node and
	edge representations learned by a pretrained molecular graph
	encoder~\cite{rong2020grover}.

	EGF stores edge states densely, so the edge tensor contains \(N^2\) entries; the largest
	graphs considered here use \(N=38\). Scaling to larger capacities therefore requires additional memory and
	pairwise computation. The reported encodings also omit stereochemistry, isotope identity,
	and radical-electron counts; supporting them would require extensions to the category sets
	and reconstruction rules. Finally, the
	reported EGF runs were based on one training seed. Replicate runs would quantify uncertainty
	in the modest margins observed in ZINC250k NSPDK-based MMD and QM9 validity.

	In proprietary molecular-generation settings, both molecular data and trained
models require protection. EGF offers a natural basis for both directions: its
continuous graph states provide a representation space for model-specific data
authorisation~\cite{wang2026catchonlyone,liu2024algo}, while the explicit separation of its
embedding, graph transformer, and readout parameters provides clear integration
points for model usage control~\cite{wang2026usagecontrol}. Future
work will evaluate these extensions and their effects on molecular validity and
generation quality.

	\section{Conclusions}
 
	EGF generates categorical graphs by transporting Gaussian states through a learned
	embedding space. Separate trainable maps define the endpoints for node and
	unordered-edge categories. A permutation-equivariant transformer predicts these
	endpoints along straight paths, and a terminal readout assigns discrete categories.
	Padding is masked, and each unordered edge follows one shared state and
	terminal decision. Across the three methods, EGF has the best value on all four
	QM9 metrics and the lowest NSPDK-based MMD on ZINC250k.
 
 
	\section*{Code and Data Availability}
	QM9 and ZINC250k are publicly available benchmark datasets
	\cite{ramakrishnan2014qm9,sterling2015zinc}. The code for training, sampling and
	evaluation, together with the configuration files for the reported runs, is
	available at \url{https://github.com/Trusted-System-Lab/EGF} under the PolyForm
	Noncommercial License 1.0.0.
 
	\small
	\bibliographystyle{plainnat}
	\bibliography{references}

\end{document}